\documentclass[review]{elsarticle}
\usepackage{lineno,hyperref}
\usepackage{multirow}
\usepackage{subfigure}
\usepackage{chngpage}
\usepackage{array}
\usepackage{booktabs}
\usepackage[american]{babel}
\usepackage{microtype}
\usepackage{amsfonts}
\usepackage{xcolor}
\modulolinenumbers[5]
\journal{Journal of Knowledge-Based Systems}
\newcommand{\revise}[1]{\textcolor{black}{#1}}
\newcommand{\tabincell}[2]{\begin{tabular}{@{}#1@{}}#2\end{tabular}}

\begin{document}
	
	\begin{frontmatter}
		
		\title{ASTRAL: Adversarial Trained LSTM-CNN for Named Entity Recognition}

		\author[address1,address2,address3]{Jiuniu Wang}\ead{wangjiuniu16@mails.ucas.ac.cn}
		\author[address1,address2]{Wenjia Xu}\ead{xuwenjia16@mails.ucas.ac.cn}
		\author[address1]{Xingyu Fu}
		\author[address1]{Guangluan Xu}
		\author[address1]{\\Yirong Wu}
		
		\address[address1]{Key Laboratory of Network Information System Technology (NIST), \\Institute of Electronics, Chinese Academy of Sciences, Beijing, China}
		\address[address2]{University of Chinese Academy of Sciences, Beijing, China}
		\address[address3]{City University of Hong Kong, Hong Kong}

		\begin{abstract}
			Named Entity Recognition (NER) is a challenging task that extracts named entities from unstructured text data, including news, articles, social comments, etc. The NER system has been studied for decades. Recently, the development of Deep Neural Networks and the progress of pre-trained word embedding have become a driving force for NER. Under such circumstances, how to make full use of the information extracted by word embedding requires more in-depth research.
            In this paper, we propose an Adversarial Trained LSTM-CNN (ASTRAL) system to improve the current NER method from both the model structure and the training process. In order to make use of the spatial information between adjacent words, Gated-CNN is introduced to fuse the information of adjacent words. Besides, a specific Adversarial training method is proposed to deal with the overfitting problem in NER. We add perturbation to variables in the network during the training process, making the variables more diverse, improving the generalization and robustness of the model. Our model is evaluated on three benchmarks, CoNLL-03, OntoNotes 5.0, and WNUT-17, achieving state-of-the-art results. Ablation study and case study also show that our system can converge faster and is less prone to overfitting.
		\end{abstract}
		
		\begin{keyword}
			\texttt{Named Entity Recognition}\sep \texttt{Deep Neural Network}\sep\texttt{Gated-CNN}  \sep\texttt{Adversarial training} 
		\end{keyword}
		
	\end{frontmatter}
	
	\section{Introduction}
	\label{sec:introduction}
	Named Entity Recognition (NER)~\cite{nadeau2007survey} is a challenging and \revise{fundamental task} in natural language processing. The NER aims to recognize named entities such as \textit{person, location, organization} from unstructured text, converting free text into the structured one. For several tasks, such as question answering and information retrieval, a NER system is often used to preprocess the data. Thus the performance of the NER would directly affect the overall performance of \revise{these advanced tasks}. Besides, scientists, especially those working on medical, biographical, and geographical, usually need to find out name entities in the literature for \revise{further} research. For example, extracting the geographic locations automatically and then displaying them on electronic maps will help people better understand and utilize the literature~\cite{zhang2009extraction}. 
	
	Over the past few years, \revise{NER} has been widely investigated.
	The development of the NER system is highly related to the evolution of the natural language processing system. 
	In the 1990s, rule-based natural language processing methods~\cite{brill1992simple, frye1995theory} prevailed, solved some easy problems. 
	However, it turns out that rule-based methods had poor versatility and are hard to transfer between domains.
	NER models could also take traditional statistic methods, such as Naive Bayes Classification~\cite{mccallum1998comparison}, CRF (Conditional Random Field)~\cite{luo2015joint} and HMM (Hidden Macov Model)~\cite{ratinov2009design}. However, these models rely on resources and features that are costly to collect.
	In recent years, deep neural networks provide a more practical solution. By learning the statistical features in a large-scale corpus, deep neural networks summarize and extract the features for specific tasks. In this paradigm, some breakthroughs appear in many tasks such as text classification, syntactic analysis, named entity recognition, information retrieval, question answering systems, etc. Furthermore, Collobert et al. proposed \revise{SENNA~\cite{collobert2011natural}}, a unified neural network architecture and learning algorithm, which can be applied to various natural language processing including NER.
	
	Recently, researchers are concerned about generating high-quality text representation, mapping natural language symbols into a high-dimensional vector space. \revise{Latest works} for text representation includes ELMo~\cite{peters2018deep}, BERT~\cite{devlin2018bert}, XLNET~\cite{yang2019xlnet}, etc. However, only improving the feature generation ability is not enough. It is an important issue to build a suitable network model and better use these text representation. BLSTM-CNN~\cite{chiu2016named} firstly combines the Bi-directional LSTM and CNN for the NER task. CNN in this model is used to extract character features and generate character embedding. Similarly, ~\cite{wang2017named} proposes CNN structure by gating mechanism, which allows more flexible information control on the CNN features. However, these methods ignore the spatial characteristic that the ``neighbor words'' can reflect the label of a certain word. For example, some words are often adjacent to the named entity, such as the articles (e.g., \textit{a, the, to}) or the verbs (e.g., \textit{love, play}). In this paper, we propose a special CNN module to process spatial features, helping to extract spatial information from adjacent words. Benefited from CNN's filter structure, the representation of each word can be closely related to the semantic information of its adjacent words. In order to control the information extracted from surrounding words, we also apply a gated mechanism \revise{within} the CNN module. 
	
	Under the stronger text representation and model structure, the performance of the NER system can be significantly improved. However, there is still a gap between the capabilities of the NER system and the industry requirements. Since the \revise{size} of NER datasets is usually not large enough, overfitting is an urgent problem for the deep neural network \revise{based NER}. So it is easy for the model to identify words that have appeared before, but hard to understand unfamiliar words. Therefore, the model needs to have a stronger generalization ability \revise{to obtain stable performance}.
	\revise{Adversarial training is a method to train the network with both the primal examples and adversarial examples. Here adversarial example means the primal example added a small adversarial perturbation which is designed to make the target model perform bad.} Adversarial training is now widely used in the image classification task, significantly increasing the generalization ability of the network against the input perturbation. For the \revise{NER} task, the input is usually discrete one-hot vectors that do not meet the infinitesimal perturbation. Instead of applying the adversarial examples to the word input, we add perturbations to the continuous word embeddings and other variables learned in the network. The adversarial examples are trained together with raw examples, improving the model's ability to withstand disturbances, and accelerating the converging process.

	We achieve a robust \revise{NER} system ASTRAL (Adversarial Trained LSTM-CNN) by augmenting the network structure and enhancing the training process. The contributions of our work are as follows:
	
	\begin{itemize}
		\item	We introduced the Gated-CNN into named entity recognition task, as an enhancement of feature extraction. We apply CNN modules on the word level, which helps the system to pay more attention to adjacent words. In order to flexibly control the spatial information extracted by CNN, we apply a gating mechanism to merge the spatial information and combine them with the original features.
		
		\item We also \revise{refine} the training process to make the NER system more stable. With adversarial training, we construct perturbations and add them to arbitrary variables in the model during \revise{each training step}, making the model have a better generalization ability. When generating perturbations, we use the target variable to constrain the norm, so that adversarial training can be applied to any variable \revise{within} the model, even to multiple variables at the same time. The experiment shows that with adversarial training, the network is \revise{much easier} to converge than the basic model.
		
		\item  We quantitatively evaluate our system on three benchmarks, which achieves the state of the art results. \revise{The experiments show} that Gated-CNN has a different influence on various types of named entities, and adversarial training is beneficial to \revise{reduce} training loss and prevent overfitting. We also perform a qualitative case study to analyze both the success and failure cases in the system. It shows the advantages of our system and the problems that need to \revise{be fixed}.
	\end{itemize}

	The remainder of this paper is organized as follows. Section 2 presents an overview of traditional and \revise{deep neural network based} methods on NER, as well as the methods for text representation and adversarial training. Section 3 describes the methodology used by our model. Section 4 verifies the effectiveness of our model by performing comparisons with the state-of-the-art methods as well as ablation experiments. Section 5 concludes the paper with discussions and outlooks.
	
	\section{Related Work}
	\subsection{Named Entity Recognition}
	Named Entity Recognition (NER) aims at detecting named entities (e.g., \textit{person, location, time,} and \textit{organization}) from unstructured text. In this \revise{subsection}, we will introduce the traditional high-performance approaches and \revise{deep neural network based} models.
    Over the last decades, numerous approaches based on traditional \revise{machine learning} algorithms are carried out on the NER task. Those methods include Naive Bayes Classifier~\cite{mccallum1998comparison}, Conditional Random Fields models (CRF)~\cite{lafferty2001conditional}, and Knowledge-driven models~\cite{wang2018label}. However, traditional methods such as Naive Bayes Classifier and Knowledge-driven models need to write too many rules according to different scenarios. Thus a specific task cannot be generalized to all the applications, making the transfer between different domains cumbersome. Besides, CRF mainly focuses on the transition probability of each word, and it does not pay \revise{enough} attention to the name entity attributes of the word.
	
	Now, most of the NER methods are based on sequence labelling~\cite{chiu2016named,graves2012supervised,lample2016neural,aguilar2019named,clark2018semi,akbik2018coling}. These methods classify every word in the corpus into different categories. These categories are corresponding to different application scenarios, such as \textit{person, location, time} and \textit{organizations}, etc. In this way, a sequence of labels which contains the entity information can be generated from these words. 
	With the \revise{developing} of deep learning techniques, the neural network has \revise{gained} state-of-the-art performance on NER. Some researchers try to reduce the manual efforts for getting labeled data. Yanyao et al.~\cite{shen2017deep} carry out incremental active learning, in which the \revise{required} amount of labeled training data can be dramatically reduced. And the lightweight architecture also speeds up the training process. \revise{These models aim to} minimize the annotation cost while maintaining the performance of NER models~\cite{chen2015study}.
	The generalization of the model is also a \revise{vital} problem worth studying. Zhenghui et al.~\cite{wang2018label} propose label-aware feature transfer learning and parameter transfer learning for cross-specialty NER. In this way, a medical NER system designed for one specialty could be conveniently applied to another one with minimal annotation efforts.
	In order to combine the advantages of previous work and get a better model ability, many researchers combine Bidirectional LSTM (Bi-LSTM)~\cite{schuster1997bidirectional} and CRF~\cite{lafferty2001conditional} to perform NER task~\cite{huang2015bidirectional,lample2016neural}. They first use Bi-LSTM to extract the text feature, then construct the CRF layer to get the output label. 
	
	\subsection{Text Representation}
	
	Text representation is a \revise{crucial} technique in natural language processing. Bengio proposed the concept of NNLM (neural network language model)~\cite{bengio2003neural} in 2003, which made the theoretical foundation for using neural networks to generate word embedding. \revise{Hence} a paradigm is formed that mapping linguistic symbols to high-dimensional spaces for further processing. After word2vec~\cite{mikolov2013efficient} and glove~\cite{pennington2014glove} are proposed, word embedding gained a better representation ability.  With large-scale corpus, the neural network based language model exerts analytical ability and achieves a lower perplexity. Since then, word embedding has become a necessary method in the field of natural language processing, performing as the representation of text in \revise{various} tasks.
	
	The text representation has great progress in recent years. There are a series of excellent works such as ELMo~\cite{peters2018deep}, GPT~\cite{radford2018improving}, BERT~\cite{devlin2018bert}, and XLNET~\cite{yang2019xlnet}. These tasks divide natural language processing into two-step: firstly use the language model to pre-train, and then use the fine-tuning module to solve various tasks. ELMo~\cite{peters2018deep} can dynamically adjust the word embedding according to the current context. GPT (Generative Pre-Training)~\cite{radford2018improving} uses Transformer~\cite{vaswani2017attention} as a feature extractor instead of RNN to obtain stronger feature extraction ability. BERT~\cite{devlin2018bert} uses the masked language model and the next sentence prediction to enhance the mining of context. XLNET~\cite{yang2019xlnet} incorporates the Transformer-XL~\cite{dai2019transformer} idea for relative segment encodings and expands the size of the dataset. These text representation methods are deeply studied in terms of pre-training, \revise{while} the construction of the application \revise{module} supporting the second stage is not focused. In this paper, instead of improving the text representation, we focus on building a better model to make use of these text representations.
	
	\subsection{Adversarial Training}
	
	Adversarial training~\cite{goodfellow2014explaining} is a method to enhance the training process with adversarial examples. Szegedy Christian et al.~\cite{szegedy2013intriguing} indicates that if the input sample is added with a well-designed perturbation, that human would not even notice, the neural network may get the wrong prediction. The sample with well-designed small perturbation is called \revise{the} adversarial example. There are two main \revise{kinds of research on} adversarial examples \revise{recently}. The first way is adversarial attacking~\cite{athalye2018obfuscated,tramèr2018ensemble}. The adversarial examples are utilized to evaluate the robustness of various models by attacking them. Additionally, the adversarial examples could be considered as extended training data to enhance the generalization and robustness of the model, which is named adversarial training. 
	
	The adversarial training method is first used on image classification task~\cite{goodfellow2014explaining}. Before updating parameters \revise{in each training step}, adversarial training examples are generated by adding perturbation to current parameters. So the adversarial training method is an augmentation of training data. Following the idea of adversarial training, Park Sungrae et al.~\cite{park2018adversarial} propose adversarial dropout by generating the mask of dropout according to the weak point of the model, which could also lead to a better training process. Adversarial training is also used in text classification~\cite{miyato2016adversarial}. In the natural language processing domain, the input of the model is discrete. So the perturbation is added to the word embedding and achieves state-of-the-art performance with a quite simple LSTM structure. After that, adversarial training is used to benefit the task of relation extraction~\cite{wu2017adversarial}. In this paper, we explore the advantage of adversarial training \revise{on the NER task}.

	\section{Methodology}
	
	\begin{figure}
		\centering
		\includegraphics[width=10cm, height=8cm]{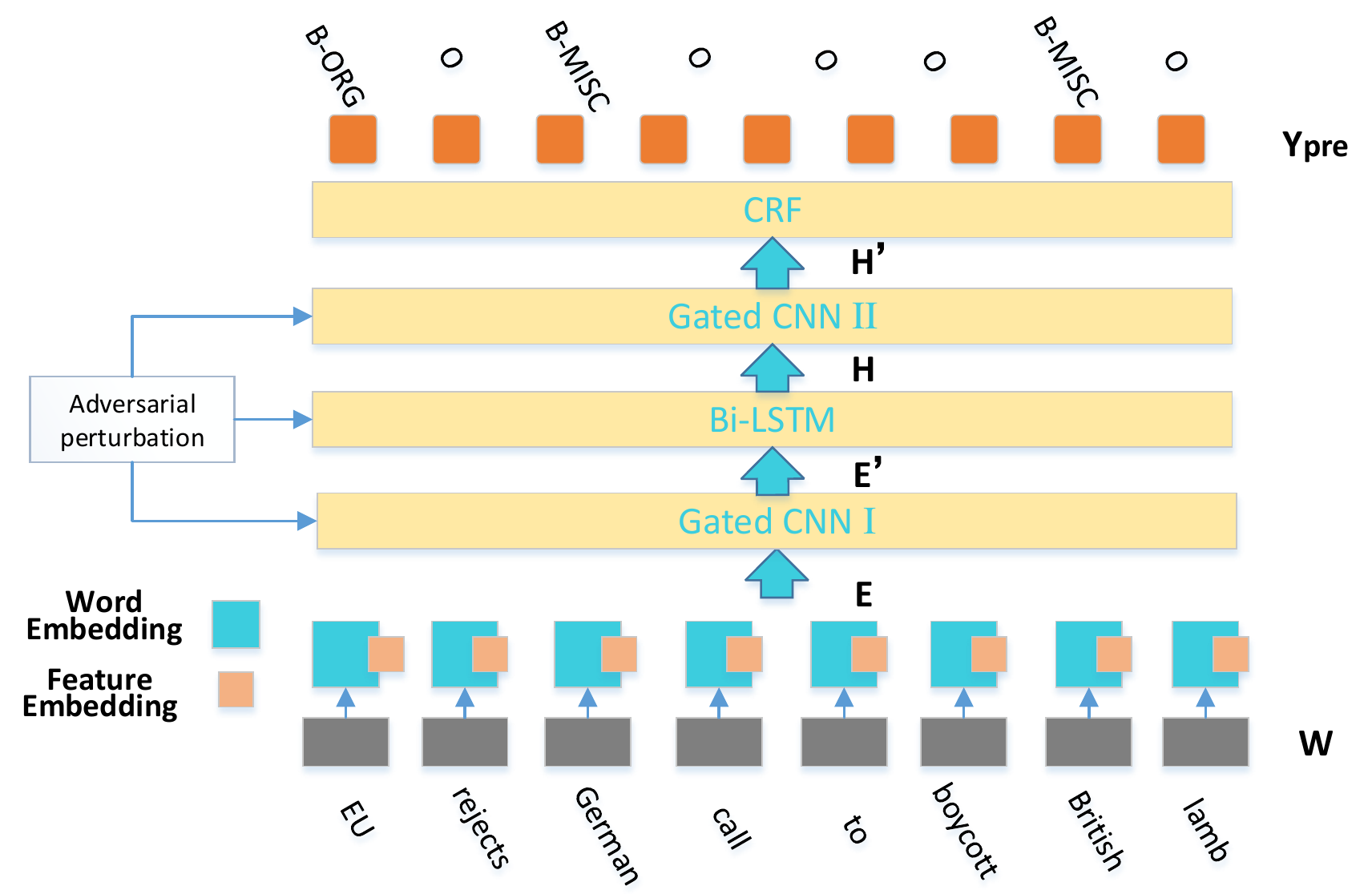}
		\caption{The overall architecture of ASTRAL. The model consists of five modules: embedding module, Gated-CNN module, Bi-LSTM module, CRF module, and adversarial training module.} \label{architecture}
	\end{figure}

	In this section, we will first demonstrate the architecture of our ASTRAL (\textbf{A}dver\textbf{S}arial \textbf{TRA}ined \textbf{L}STM-CNN) model, then illustrate the implementation detail of adversarial training.
	
	The overall structure of ASTRAL is illustrated in Figure~\ref{architecture}. 
	As shown in this figure, the goal of ASTRAL is to \revise{predict} tags $Y_{pre}$ with the same length of the input sentence $W$. Here $W=({{w}_{1}},{{w}_{2}},\dots,{{w}_{n}})$ represents a sentence with $n$ tokens, ${{Y}_{pre}}=({{y}_{1}},{{y}_{2}},\dots,{{y}_{n}})$ represents $n$ predicted tags for tokens in $W$. In our model, IOB format (short for \textit{inside, outside,} and \textit{beginning}) is used as the label standard. Since there are multiple types of named entities, suffixes are attached to represent their entity type after the B and I. So the tag in ${Y}_{pre}$ could be B-\#, where \# is related to the specific named entity type, \revise{e.g., ORG, MISC}. For example, in Figure~\ref{architecture}, when identifying the sentence ``EU rejects German call to boycott British lamb'', we can determine that ``EU'' belongs to organization (ORG), while ``German'' and ``British'' belong to miscellaneous (MISC), thus the sequence of tags would be ``B-ORG, O, B-MISC, O, O, O, B-MISC, O''.
	
	The ASTRAL model \revise{is composed of} five modules: embedding module, Gated-CNN module, Bi-LSTM module, CRF module, and adversarial training module. Embedding module transforms the words into vectors. Bi-LSTM module is a variant of RNN (Recurrent Neural Network), which generate features from word vectors. CNN can enhance the refine of spatial features, and the gate mechanism further filters the obtained information. The CRF module combines the information acquired by the Bi-LSTM and the Gated-CNN, then \revise{generates} the final \revise{tags as the} output. 
    During training, the adversarial training module generates adversarial perturbation to make the model more generalized and \revise{obtain} better training accuracy.
	
	\subsection{Embedding Module}
	\revise{
	Given a sentence $W=(w_1,w_2,\dots,w_n)$ with $n$ tokens, the embedding module aims at transferring $W\in {\mathbb{R}^{{n_{id}} \times n}}$ into its embedding representation $E=(e_1,e_2,\dots,e_n)$, where $w_i\in{{\mathbb{Z}}^ +}$ denotes the index of the $i$-th token in the sentence, $e_i\in \mathbb{R}^{d_e}$ corresponds to the $i$-th token, $n_{id}$ is the number of all used tokens. In our model, $E \in {\mathbb{R}^{{d_e} \times n}}$ is the concatenation of ${E}_{w}$ and ${E}_{f}$ as
	\begin{equation}
	E=[{{E}_{w}};{{E}_{f}}] \,, \\ 
	\end{equation}
	where $[\cdot;\cdot]$ denotes the concatenation of different vectors, ${E_w} \in {\mathbb{R}^{{d_w} \times n}}$ denotes the pooled contextualized embedding~\cite{akbik2019pooled}, ${E_f} \in {\mathbb{R}^{{d_f} \times n}}$ denotes the feature embedding, ${d_e} = {d_w} + {d_f}$, ${d}_{w}=300$ and ${d}_{f}=20$ in our experiments.
	We then introduce the definition and function of these two submodules in detail. Pooled contextualized embedding~\cite{akbik2019pooled} ${E}_{w}$ is a kind of general word embedding
	\begin{equation}
	{{E}_{w}}={M}_{w}\cdot W \,,
	\end{equation}
	where ${M}_{w} \in {{\mathbb{R}}^{{{d}_{w}}\times {{n}_{id}}}}$ denotes the matric of pre-trained pooled contextualized embedding. ${E}_{w}$ contains contextual meaning around the target word and previous memory meaning appeared in the dataset before. Contextualized embedding can produce meaningful embeddings for even rare string by using the memory mechanism instances. And pooling operation helps to distill word representation from all contextualized tokens.
	Then we utilize feature embedding ${{E}_{f}}$ to extract rule-based information
	\begin{equation}
	{{E}_{f}}={M}_{f}\cdot W_f \,,
	\end{equation}
	where ${M}_{f} \in {{\mathbb{R}}^{{{d}_{f}}\times {{n}_{f}}}}$ denotes the parameter matric of feature embedding, and ${W_f} \in {\mathbb{R}^{{n_f} \times n}}$ denotes the features indicator of given tokens. The capitalization of words is obviously useful when discriminating  named entities, e.g., a location usually starts with an upper character. So following the previous work~\cite{ghaddar2018robust}, our selected five features are all-lower, upper-first, upper-not-first, numeric, and no-alpha-num, which means ${n_f=5}$. Then the sentence feature $W_f$ is mapped by the random initialized lookup table ${M}_{f}$ to  ${E}_{f}\in {\mathbb{R}}^{{{d}_{f}}\times {n}}$ which contains $n$ vectors with $d_f$ dimension. After training, feature embedding ${E}_{f}$ can establish an effective representation relationship with named entities.
	}
	
	\subsection{Gated-CNN Module}
	\begin{figure}
		\centering
		\includegraphics[width=7.5cm, height=7cm]{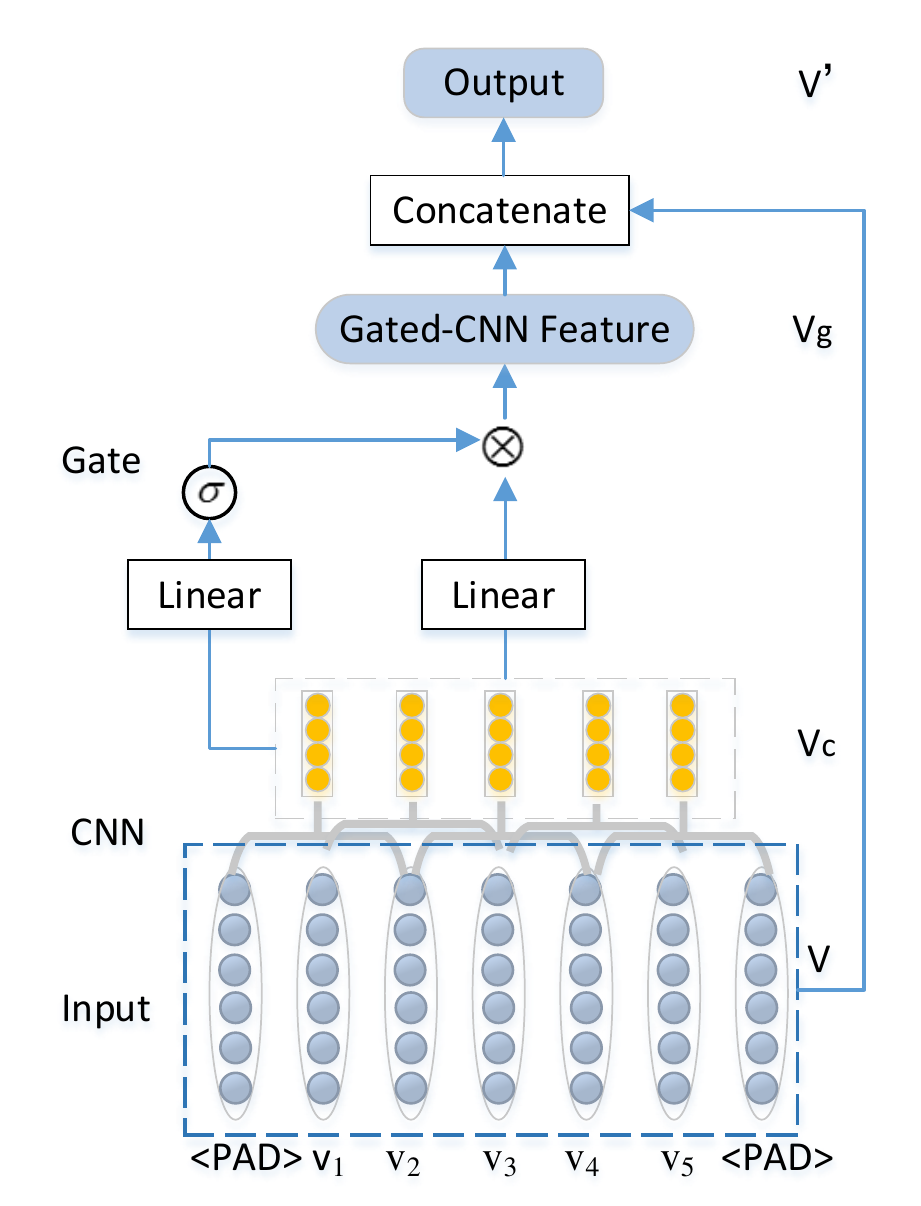}
		\caption{The structure of Gated-CNN. $V$ denotes the input variable which could be embedding $E$ or hidden states of Bi-LSTM $H$. The input variable $V$ passes a CNN with the filter size $3*N$ and get the feature containing spatial information (represented in yellow). Then two linear functions are used to get the Gated-CNN feature $V_g$.} \label{fig_Gated_CNN}
	\end{figure}
	In this model, the Gated-CNN module is proposed to integrate the spatial information extracted by the \revise{adjacent} words. The structure of the Gated-CNN module is shown in Figure~\ref{fig_Gated_CNN}, which consists of one CNN and two linear layers. \revise{Given} the input sentence variable with $n$ tokens $V=({{v}_{1}},{{v}_{2}},...,{{v}_{n}})$, we first calculate the integrated representation for each token with its adjacent tokens:
    \begin{equation}
	{{V}_{c}}={{f}_{CNN}}(V)
	\end{equation}
	where ${{f}_{CNN}}(\cdot)$ denotes the function of CNN. This is achieved by one filter with a size of $N_{w}\times N_{o}$, where window size $N_{w}$ is set in [3,5,7], meaning the number of tokens that are processed at a time and $N_{o}$ is a hyperparameter related to the output vector size. So the feature vector of each token is related to its adjacent tokens. Under the effect of padding, each column of the vector ${V}_{c} =({{v}_{c1}},{{v}_{c2}},...,{{v}_{cn}}) $ obtained by CNN can still correspond to the original token. Therefore, the vector representation of the $i$-th token ${v}_{ci}$ synthesizes the spatial information of its two \revise{sides'} surrounding words.
	
	Then a gated linear layer is proposed to control the feature vectors produced by the CNN layer: 
	\begin{equation}
	{{V}_{g}}=({{W}_{1}}\cdot {{V}_{c}}+{{b}_{1}})\otimes \sigma ({{W}_{2}}\cdot {{V}_{c}}+{{b}_{2}}) \\ 
	\end{equation}
	where ${W}_{1}$, ${W}_{2}$, ${b}_{1}$, ${b}_{2}$ are training parameters of linear functions, $\otimes$ denotes element-wise product, and $\sigma$ denotes the sigmoid function. The gate is trained through the dataset, and it roughly decreases the task-independent vectors to reduce the noise, while amplifying the task-related vectors to enhance the network focus. The gate makes the variables more responsive to the task by changing the focus on the feature map $V_c$.
	
	Finally, we concatenate the variable ${V}_{g}$ with $V$, integrating spatial information and the original information to \revise{get a} more vibrant text representation $V'$ as
	\begin{equation}
	V'=[V;{{V}_{g}}] \,.
	\end{equation}
	
	In this model, the Gated-CNN module is used twice, one for embedding and the other for contextual extraction. As it is shown in Figure~\ref{architecture}, for Gated-CNN I, the input variable $E$ is the embedding representation of the sentence, and we get $E'=G(E)$. For Gated-CNN II, the integrated high-level variable $H$ is processed. It is the same for $H'=G(H)$ when Gated-CNN is used for the hidden state variable of Bi-LSTM $H$.

	\subsection{Bi-LSTM Module}
	\begin{figure}
		\centering
		\includegraphics[width=7.5cm, height=6cm]{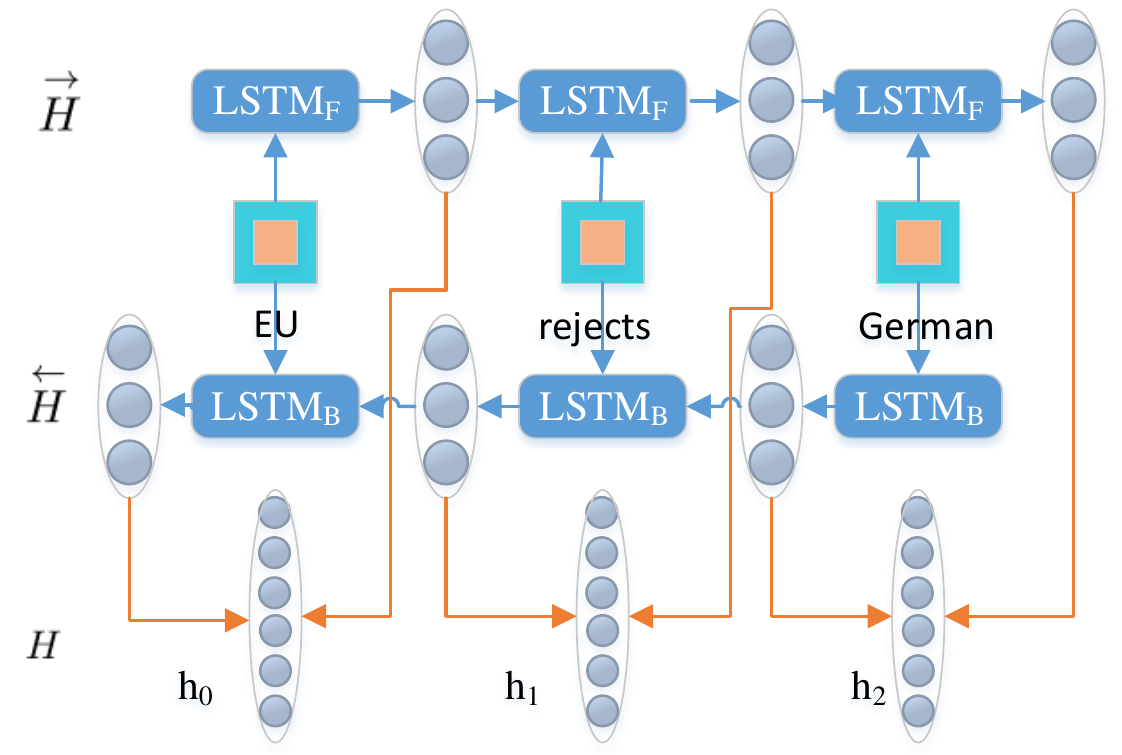}
		\caption{The structure of Bi-LSTM. ${LSTM}_F$ is the forward LSTM, ${LSTM}_B$ is the backward LSTM. The output of Bi-LSTM $H$ is the concatenation of these two sub LSTMs' output.} \label{fig2}
	\end{figure}
	
	LSTM (Long Short Term Memory)~\cite{iet} is a kind of RNN (Recurrent neural network), which extracts the features in the chronological order of the input.  And the formulation of Bi-LSTM can be described as:
	\begin{equation}
	\begin{array}{l}
	\stackrel{\rightarrow}{H}=LST{{M}_{F}}(E') \\
	\stackrel{\leftarrow}{H}=LST{{M}_{B}}(E') \\ 
	H=[\stackrel{\rightarrow}{H};\stackrel{\leftarrow}{H}] \,. \\ 
	\end{array}
	\end{equation}
	In this paper, we use Bi-LSTM (Bidirectional LSTM) to extracts the features from both \revise{forward} direction as $\stackrel{\rightarrow}{H}$ and \revise{backward} direction as $\stackrel{\leftarrow}{H}$. The network structure is shown in Figure~\ref{fig2}. \revise{It obtains the representation of each token in turn from both the forward and the backward directions, finding out the correlation between other surrounding words.}
	
	\subsection{CRF Module}
	The use of CRF (Conditional Random Field) in conjunction with Bi-LSTM is a standard method for the sequence labeling task. As shown in Figure~\ref{architecture}, the input variable of CRF is  $H'$ generated by Gated-CNN II, and its output is predicted tags $Y_{pre}=({{y}_{1}},{{y}_{2}},...,{{y}_{n}})$.
    CRF generates sequence tags $Y_{pre}$ by status feature function $s_k(y_i,H',i)$ and the transition feature function $t_j(y_{i+1},y_i,H',i)$. And the $s_k(y_i,H',i)$ indicates the influence of the input variable $H'$ on $y_i$. The $t_j(y_{i+1},y_i,H',i)$  depicts the effect of $H'$ on the adjacent tag changes in $Y_{pre}$. The predicted tags $Y_{pre}$ is generated by maximum the score
	\begin{equation}
	 P(y|x)=\frac{1}{Z}\exp (\sum\limits_{j}{\sum\limits_{i=1}^{n-1}{{{\lambda }_{i}}{{t}_{j}}({{y}_{i+1}},{{y}_{i}},H',i)}+\sum\limits_{k}{\sum\limits_{i=1}^{n}{{{\mu }_{k}}{{s}_{k}}({{y}_{i}},H',i)}}}) \,,
	\end{equation}
	where ${\lambda }_{i}$ and ${\mu }_{k}$ are hyperparameters, and $Z$ is the normalization factor. The CRF module can learn the constraints of the sequence tags. For example, the beginning of a sentence should be ``B'' or ``O'' instead of ``I''. ``O I'' is impossible since the beginning of the named entity should be ``B'' instead of ``I''.

	\subsection{Adversarial Training Module}

	In general, the purpose of the deep neural network is to get predicted output ${Y_{pre}}$ by the input ${V_{in}}$, making the predicted result ${Y_{pre}}$ and the ground truth ${Y}$ closer. The model learns the parameters $\theta$ to minimize the loss function
	\begin{equation}
	L = loss({Y_{pre}},Y) \,,
	\end{equation}
	where commonly used loss function includes L1Loss, MSELoss (mean squared error), CrossEntropyLoss, NLLLoss (Negative Log Likelihood), etc. We use CrossEntropyLoss in our experiments.
	
\begin{figure}
		\centering
		\includegraphics[width=12cm, height=3.5cm]{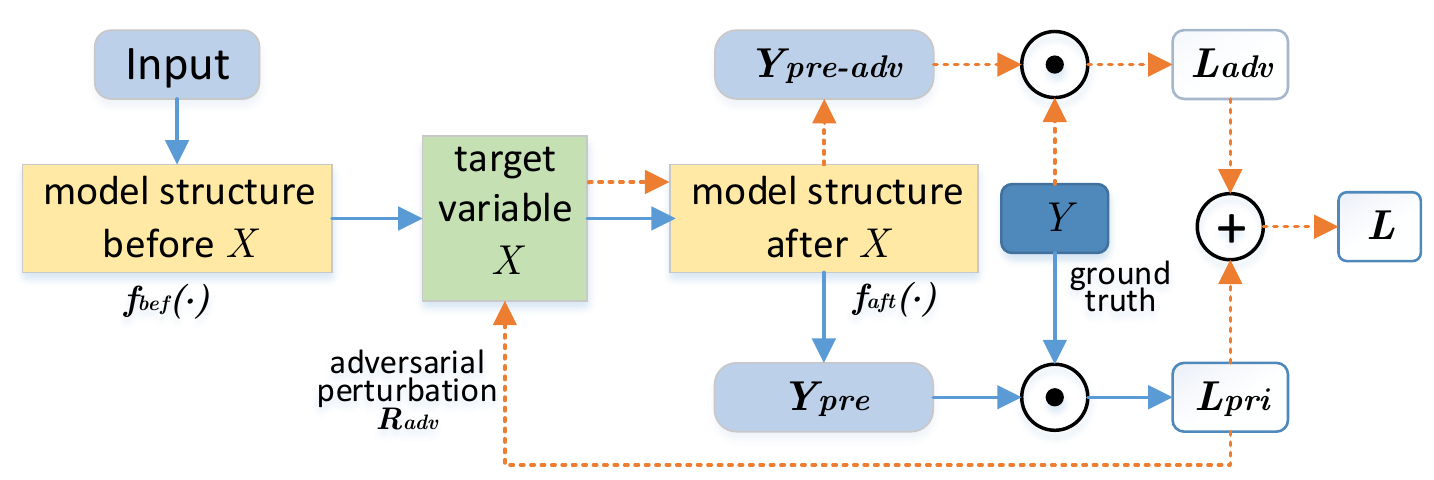}
		\caption{The flowchart of adversarial training. The solid line (blue) shows the first round process of obtaining \revise{primal} loss $L_{pri}$. The dashed line (orange) shows the second round process of calculating $R_{adv}$ according to $L$ and $X$, further obtaining $L_{adv}$, and then finally generating final loss $L$. Here $\odot$ denotes the loss function, $\oplus $ denotes add operation; $Y_{pre}$ and $Y_{pre-adv}$ represent prediction results with and without adversarial perturbation respectively. The final optimized loss $L$ is the sum of \revise{primal} loss $L_{pri}$ and adversarial loss $L_{adv}$.} \label{adversarial}
	\end{figure}

	In this section, we describe how to use normalized adversarial training to strengthen the training process. As shown in Figure~\ref{adversarial}, for every variable $X$ in the model, we can regard it as the adversarial training target variable and add perturbation on it. We represent the model before $X$ as ${f_{bef}}(\cdot)$, and the model after $X$ as ${f_{aft}}(\cdot)$. In our model, we choose the output of Gated-CNN modules $E'$ and $H'$ as the target variables.
	
	The adversarial training process in our model can be divided into two rounds. In the first round, our model generates \revise{primal} loss $L_{pri}$ based on the input. 
	\begin{equation}
	X = {f_{bef}}({V_{in}};\theta ),
	{Y_{pre}} = {f_{aft}}(X;\theta) \,,
	\end{equation}
	where $V_{in}$ is the input variable for the model. And the \revise{primal} loss is 
	\begin{equation}
	L_{pri} = F(X,Y;\theta ) = loss({f_{aft}}(X;\theta ),Y) \,.
	\end{equation}
	\revise{
	In the second round, $L_{pri}$ is derived from $X$ and normalized to obtain adversarial perturbation $r_{adv}$. 
	Here ${r_{adv}}$ should theoretically be obtained from the following optimization problems:
	\begin{equation}
	{r_{adv}} = \mathop {\arg \max }\limits_{r,||r|| \le \varepsilon } F(X + r,Y;\hat \theta )\,,
	\end{equation}
	where $\varepsilon $ constraints the norm of ${r_{adv}}$, and $\hat \theta$ indicates the instantaneous value of the parameter for each solution. The parameters are constantly updated, thus the value of $\hat \theta $ is different for each training sample and training step.	
	In order to get the numerical solution for ${r_{adv}}$, we apply an approximate solution~\cite{goodfellow2014explaining}. The $F(X,Y;\hat \theta )$ is assumed as a linear function around $X$, so the approximated value of ${r_{adv}}$ can be defined as:
	\begin{equation}
	{r_{adv}} = \varepsilon X \otimes d/||d||,
	d = {\nabla _X}F(X,Y;\hat \theta )\,,
	\end{equation}
	where $d$ is the gradient of the primal loss $\frac{{\partial {L_{pri}}}}{{\partial X}}$, $\varepsilon$ is a hyperparameter, $\otimes$ denotes element-wise product and ${r_{adv}}$ is the adversarial perturbation designed to ascend the current loss. $X$ is introduced as the multiplicator when calculating ${r_{adv}}$, because it is more robust when simultaneously using ${r_{adv}}$ of multiple target variables under such normalization.
	Then the sum of $r_{adv}$ and $X$ is put into the ${f_{aft}}(\cdot)$ (structure after $X$) to get adversarial loss $L_{adv}$ as
	\begin{equation}
	L_{adv} = loss({f_{aft}}(X + {r_{adv}};\theta ),Y)\,.
	\end{equation}
	The final optimized loss is the sum of these two losses as
	\begin{equation}
	L = L_{pri} + L_{adv} \,.
	\end{equation}
	The model parameters $\theta$ optimized in this way can be adapted to both the original data and the disturbing data.
	}
	
	\section{Experiments and Results}
	\subsection{Dataset and Criteria}
	\subsubsection{Dataset}
	
	In this paper, we apply our NER system to three English datasets, CoNLL-03~\cite{sang2003introduction}, OntoNotes 5.0~\cite{pradhan2012conll} and WNUT-17~\cite{derczynski2017results}, showcasing the effectiveness and robustness of our system. CoNLL-03~\cite{sang2003introduction} is a large dataset widely used by NER researchers, whose data source is Reuters RCV1 corpus, leading its main content to be newswire. Its named entities include location, organization, person, and miscellaneous. OntoNotes 5.0~\cite{pradhan2012conll} is a larger dataset which was initially built for CoNLL 2012 shared task. The source of the text in the dataset was the LDC2013T19~\cite{weischedel2013ontonotes} published by the Linguistic Data Consortium. It covers a wide range of content, including telephone conversations, newswire, newsgroups, broadcast news, broadcast conversation, and weblogs. WNUT-17~\cite{derczynski2017results} is a complex dataset from various sources, which is mainly derived from social media. The training set is extracted from tweets, while the development set comes from the comments of YouTube, and the testing set is based on Reddit and StackExchange. The inconsistent data for training and testing make it difficult to recognize named entities for WNUT-17.

	\begin{table}
		\caption{Dataset statistics. The size of datasets is in the number of entities/tokens.}\label{dataset_statistic}
		\centering
		\footnotesize
		\resizebox{\textwidth}{12mm}{
			\begin{tabular}{|c|c|c|c|c|c|}
				\hline
				\textbf{Dataset} & \textbf{Train} & \textbf{Dev} & \textbf{Test} & \tabincell{c}{\textbf{Entities}\\\textbf{Frequency}}	& \tabincell{c}{\textbf{Entity}\\\textbf{Types}} \\
				\hline
				CoNLL-03 &  23,499 / 204,567  &5,942 / 51,578&5,648 / 46,666& 11.6\% & 4\\
				\hline
				OntoNotes 5.0 & 81,828 / 1,088,503&11,066 / 147,724&11,257 / 152,728& 7.5\% & 18\\
				\hline
				WNUT-17 & 3,160 / 62,729& 1,250 / 15,733& 1,589 / 23,394& 5.9\% & 6\\
				\hline
			\end{tabular}
		}
	\end{table}
	
	\begin{figure}
		\centering
		\includegraphics[width=7cm, height=5cm]{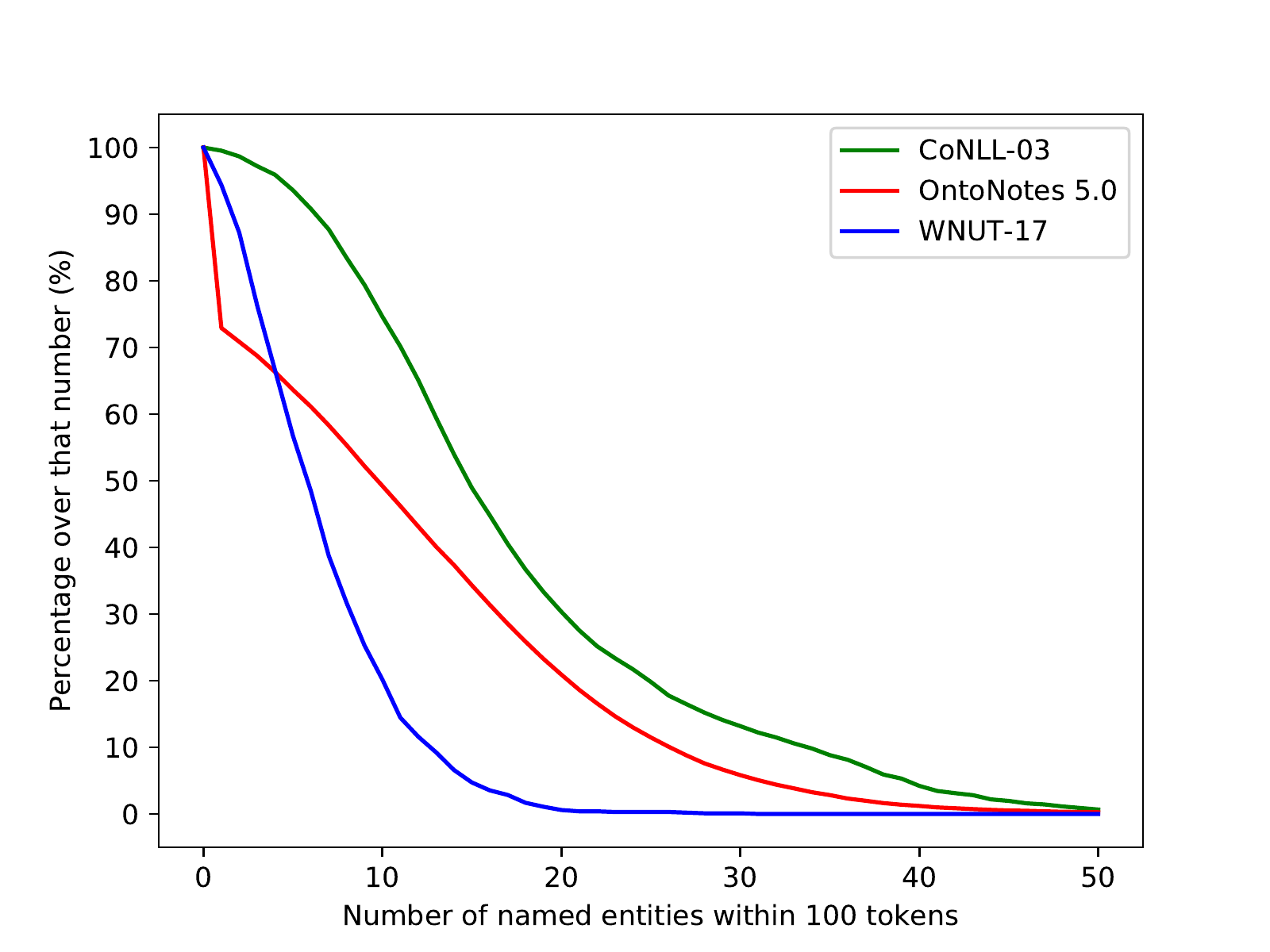}
		\caption{\revise{The distribution of named entities on the three datasets. The number of named entities within every 100 tokens is counted, and we show the percentage over that number on each dataset. We only show the number from 0 to 50 since few cases with more than 50 named entity tokens within 100 tokens.}} \label{fig_distribution_of_dataset}
	\end{figure}

	We show the statistics of the above datasets in Table~\ref{dataset_statistic}. When evaluating the NER system, \revise{researchers} are more inclined to compare their results on CoNLL-03. From Table~\ref{dataset_statistic}, we can see that the token and entity size of OntoNotes 5.0 is the largest, which helps to test the generalization ability of our network on large datasets. WNUT-17, a dataset closer to daily lives, makes more sense for the practical implication of the NER systems. \revise{We also analyse the distribution of named entities by the column ``Entities Frequency'' in Table~\ref{dataset_statistic} and the curves in Figure~\ref{fig_distribution_of_dataset}. The frequency of entities for the three datasets is quite different. 11.6\% of tokens in CoNLL-03 are named entities, while only 5.9\% of that in WNUT-17. Figure~\ref{fig_distribution_of_dataset} specifically indicates this phenomenon. We divide every 100 tokens into a group, and the percentage in CoNLL-03 that contains ten or more entity tokens is 70\%, while that in WNUT-17 is only 14\%. It means the percentage of entity tokens in WNUT-17 is relatively small.}

	\subsubsection{Evaluation Metrics}
	
	In the experiment, we mainly measure the F1 values of different models in the above three datasets.
	Precision ($P$), Recall ($R$), and $F1$ value are common indicators for measuring model performance:
	\begin{equation}
	P = \frac{{|A|}}{{|{T_{pre}}|}},R = \frac{{|A|}}{{|{T_{gt}}|}},F1 = \frac{{2PR}}{{P + R}},
	\end{equation}
	where ${T_{pre}}$ represents the predicted answer collection, ${T_{gt}}$ denotes the ground truth answer collection, $A = {T_{pre}} \cap {T_{gt}}$ is the hit answers, and $|\cdot|$  is the number of elements in the collection.
	In detail, we measure the performance of the system for each word. For example, as a named entity consisting of two words with labels ``B-PER I-PER'', it is considered to be two essential elements when evaluating.
	
	\subsection{Main Results}
	
	\begin{table*} 
		\caption{Test F1 score for different models on the datasets. \revise{In this table, ``$^*$'' indicates the results implemented by us, and ``-'' indicates that the performances of the models on the corresponding datasets are not yet obtained.} }\label{tab_main_results}
		\centering
		\resizebox{\textwidth}{22mm}{
			\begin{tabular}{|c|c|c|c|}
				\hline
				\textbf{Model} & \textbf{CoNLL-03}  & \textbf{OntoNote 5.0} & \textbf{WNUT-17}\\
				\hline
				Character-LSTM~\cite{lample2016neural} & 90.94 & \revise{84.86$^*$} & \revise{44.79$^*$}  \\
				\hline
				BLSTM-CNN~\cite{chiu2016named} & 91.62 & 86.28 & \revise{45.14$^*$} \\
				\hline
				Stacked Multitask~\cite{aguilar2019named} & - & -  & 45.55 \\
				\hline
				ELMo~\cite{peters2018deep} & 92.22 & - & - \\
				\hline
				CVT+Multitask~\cite{clark2018semi} & 92.6 &-  & - \\
				\hline
				BERT~\cite{devlin2018bert} & 92.81 & \revise{88.28$^*$} & \revise{49.23$^*$}  \\
				\hline
				Contextual String Embedding~\cite{akbik2018coling}	 & 92.86 & 88.75 & 49.49 \\
				\hline
				ASTRAL (ours) & \textbf{93.32} & \textbf{89.44} & \textbf{49.72} \\
				\hline
			\end{tabular}
		}
	\end{table*}
	
	We perform experiments on three datasets, CoNLL-03, OntoNote 5.0, and WNUT-17, to measure the models' ability to identify named entities. The tested models include those focus on model improvements, such as Character-LSTM~\cite{lample2016neural} and BLSTM-CNN~\cite{chiu2016named}, and those focus on word embedding and representation, such as ELMo~\cite{peters2018deep} and BERT~\cite{devlin2018bert}. 
	The quantitative results of our model are shown in Table~\ref{tab_main_results}. Since CoNLL-03 is widely used by most of the models, the experimental results of former research are sufficient, which is also the most convincing measure of system performance. \revise{In order to strengthen the integrity of the experiment, we implement several models, i.e., Character-LSTM, BLSTM-CNN, and BERT. And these implemented results are marked with ``$^*$''  in Table~\ref{tab_main_results}. Although some other complex models still lack some results which are marked with ``-'', we believe that the current results are sufficient for experimental analysis.} Before the methods with pre-training language models such as ELMo~\cite{peters2018deep}, the model could not achieve 92\% in CoNLL-03. While with the language model like ELMo~\cite{peters2018deep}, BERT~\cite{devlin2018bert}, and other large-scale pre-training methods, the performance of the model has been significantly improved up to 92.81\%. Our model follows the language model method, focusing on improving the model structure and training method. It can achieve 93.32\% F1 on the CoNLL-03. \revise{The improvement can also be found on both OntoNote 5.0 and WNUT-17 by improving the model structure or the word representation. Especially on WNUT-17 dataset, the BERT model has a 3.68\% improvement over Stacked Multitask. It shows that the pre-training language model benefits more on the dataset with the complex and diverse language.} Our model also performs well on more complex datasets OntoNote 5.0 and WNUT-17. The experimental results show that ASTRAL has got state-of-the-art results in the NER task.

	\subsection{Effect of Model Architecture}
	\paragraph{Ablation Study}
	
	In order to verify the validity \revise{of our modules}, we conducted an ablation study. As it is shown in Table~\ref{tab_ablation_study}, we conducted experiments on the four conditions of ASTRAL for three datasets. Here $Basic$ indicates the basic model with pre-trained word embedding and Bi-LSTM. $GC$ indicates that only the Gated-CNN is added to the basic model. $AT$ indicates that only the adversarial training method is added to the basic model. $ATGC$ indicates that the complete ASTRAL model includes Gated-CNN and adversarial training. 
	As can be seen from the results in Table~\ref{tab_ablation_study}, Gated-CNN and adversarial training both benefit the overall results. Finally, the combination of Gated-CNN and adversarial training can achieve better experimental results. It causes accuracy increase for 0.42\% on CoNLL-03 dataset, 0.67\% on OntoNote 5.0 dataset, and 0.57\% on WNUT-17 dataset respectively.
    \begin{table} 
		\caption{Ablation study for our ASTRAL model. Here ``Basic'' denotes basic model, ``GC'' denotes Gated-CNN, ``AT'' denotes Adversarial Training, and ``ATGC'' denotes the combination of GC and AT.}\label{tab_ablation_study}
		\centering
		\footnotesize
			\begin{tabular}{|c|c|c|c|c|}
				\hline
				\multicolumn{2}{|c|}{\textbf{Model}} & \textbf{CoNLL-03}  & \textbf{OntoNote 5.0} & \textbf{WNUT-17}\\
				\hline
				\multirow{4}*{ASTRAL} & Basic& 92.92 & 88.77 & 49.15  \\
				\cline{2-5}
				&GC & 93.04 & 89.02 & 49.38 \\
				\cline{2-5}
				&AT & 93.18 & 89.23 & 49.65 \\
				\cline{2-5}
				&ATGC & \textbf{93.32} & \textbf{89.44} & \textbf{49.72} \\
				\hline
			\end{tabular}
	\end{table}
	
	\begin{figure}
		\centering  
		\subfigure[CoNLL-03]{
			\label{CoNLL-03-score}
			\includegraphics[width=0.8\textwidth]{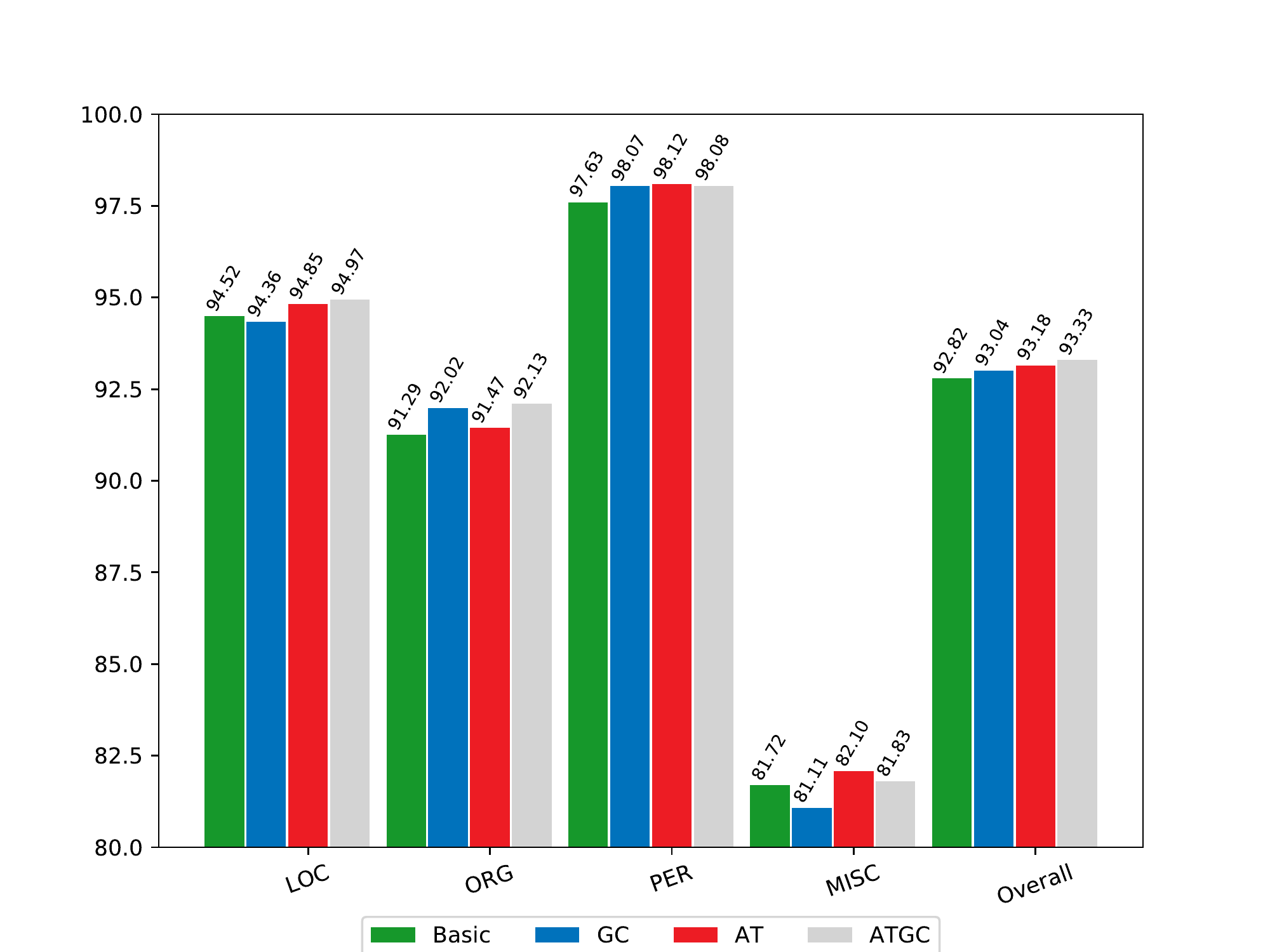}}
		\subfigure[WNUT-17]{
			\label{WNUT-17-score}
			\includegraphics[width=0.8\textwidth]{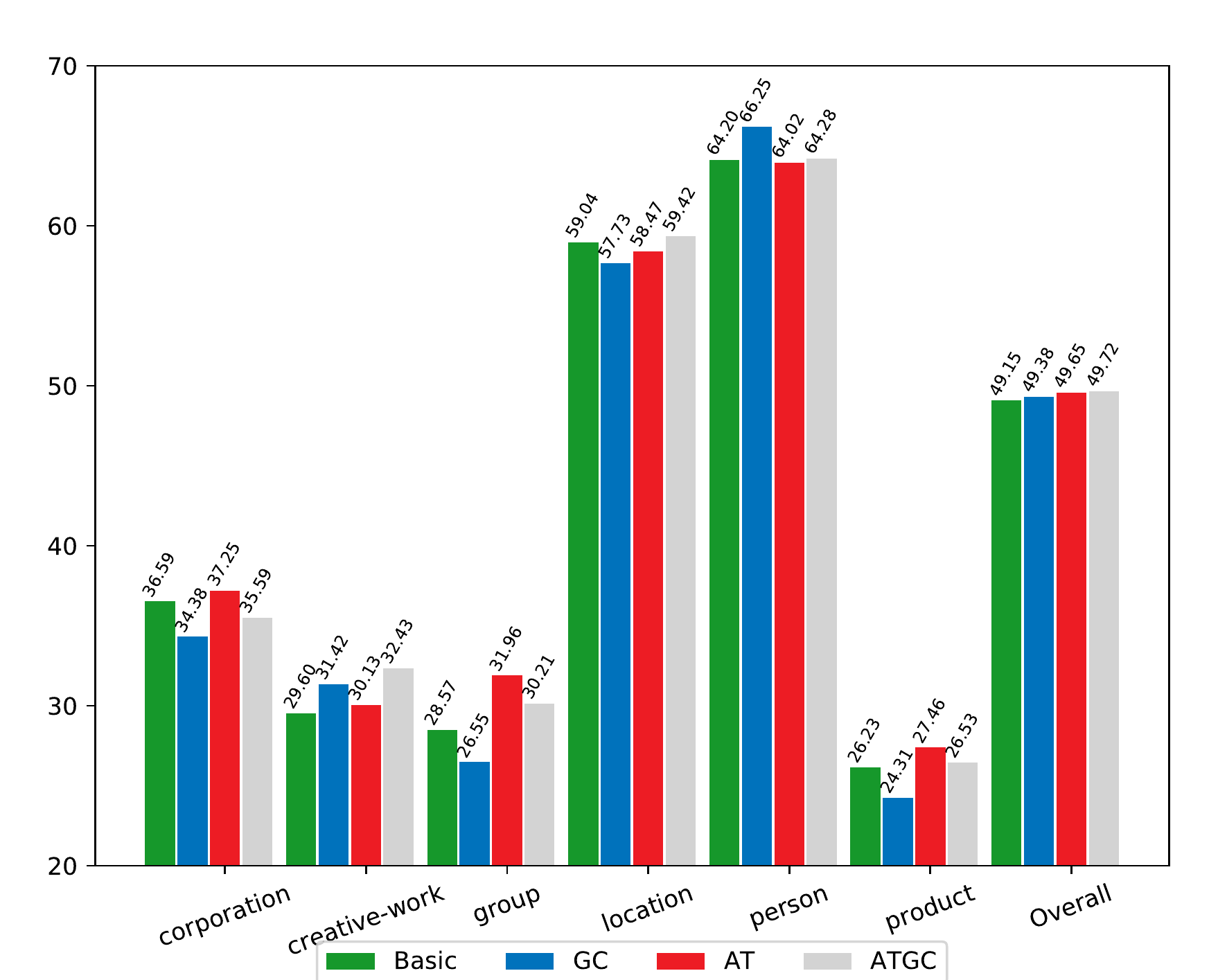}}
		\caption{The impact of changes in the model structure on various entity types. The results on CoNLL-03 dataset and WNUT-17 dataset are shown in the figure. Here ``Basic'' denotes basic model, ``GC'' denotes Gated-CNN, ``AT'' denotes Adversarial Training, and ``ATGC'' denotes the combination of GC and AT.}
		\label{accuracy_for_different_entity}
	\end{figure}	
	
	Figure~\ref{accuracy_for_different_entity} shows the \revise{model performance on} different entity types in the two datasets CoNLL-03 and WNUT-17. When the model structure changes, the specific F1 values of different entity types are also different. Gated-CNN leads a significant improvement on the ORG (organization), PER (person) in CoNLL-03, \revise{as well as} the creative-work, person in WNUT-17. One reasonable explanation lies that the Gated-CNN emphasizes the attention of each word to its \revise{adjacent} words, and there are usually specific words (such as ``at'', ``to'', etc.) around these benefited named entities. But it has little or even adverse effect on certain entity types such as corporation and product on WNUT-17, which indicates that the adjacent words might have a negative impact on recognizing some kinds of entities. If the relationship between \revise{adjacent} words and named entities is not obvious, then Gated-CNN will bring some noise to the system. Unlike Gated-CNN, adversarial training has improved the performance on almost all kinds of entities, indicating its stability.
	
	\paragraph{Model Generalization}
	
	\begin{figure} 
		\centering
		\includegraphics[width=9.5cm, height=6cm]{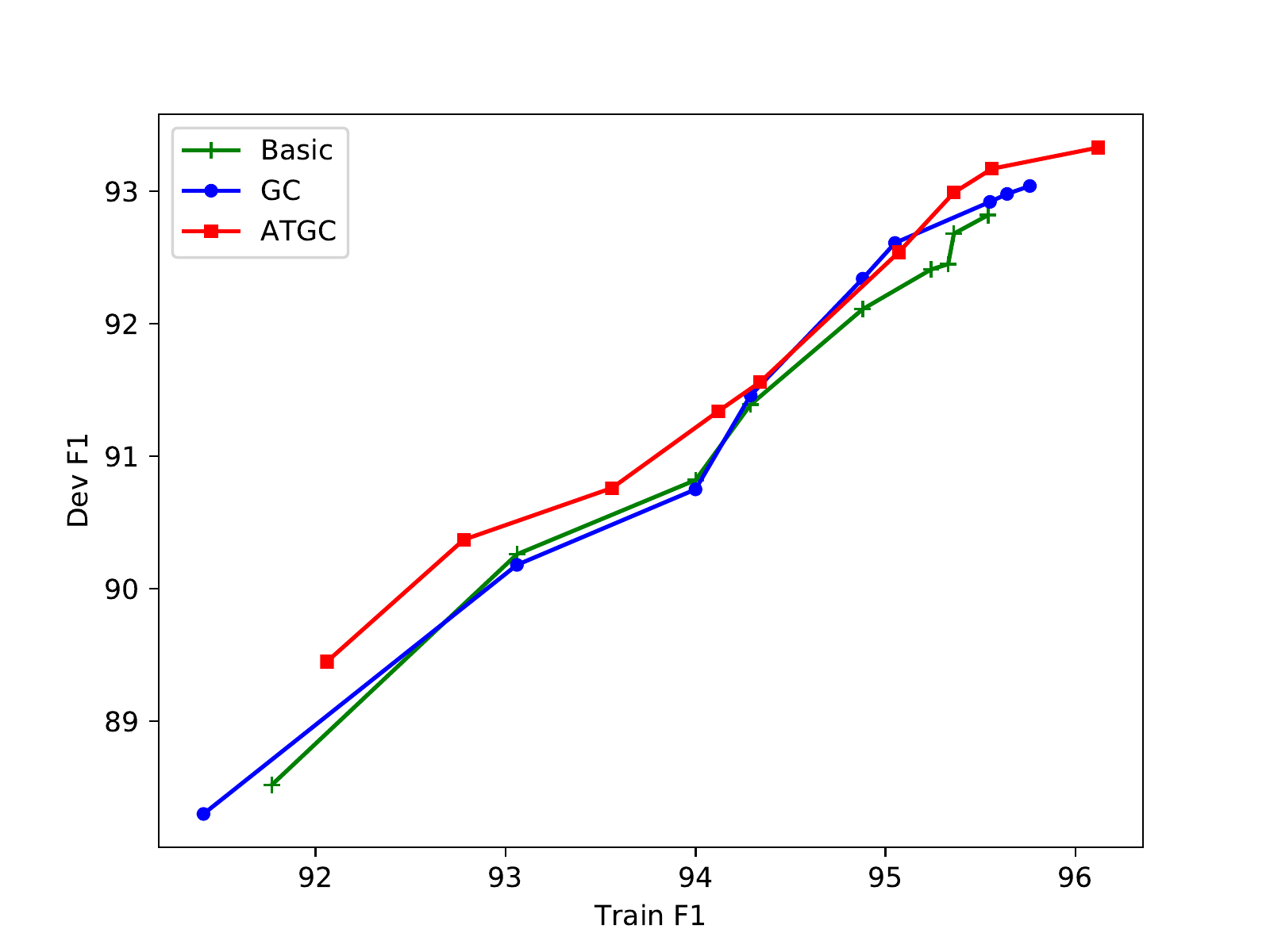}
		\caption{Dev F1 - Train F1 curves for different model conditions. Here ``Basic'' denotes basic model, ``GC'' denotes Gated-CNN, and ``ATGC'' denotes the combination of GC and AT.} \label{fig_generalization}
	\end{figure}

	Figure~\ref{fig_generalization} shows the Dev F1 - Train F1 curve under three conditions: basic model (the green curve), GC (the blue curve), and ATGC (the red curve). Dev F1 and Train F1 indicate the model performance on validating set and training set respectively in the training process.

	The curves of \revise{$Basic$ (green) and $GC$ (blue)} in Figure~\ref{fig_generalization} is close, indicating \revise{our basic model and Gated-CNN model} have similar generalization ability. The position of the red curve is on the upper side of the other curves, indicating that the Dev F1 value of ATGC is higher under the same Train F1. So we can conclude that ATGC model has better generalization ability. 
	Additionally, observing the upper right corner of Figure~\ref{fig_generalization}, it \revise{is obvious} that Basic, GC, and ATGC can reach upper and upper positions, respectively. It shows that the training level of the model is deepened in these three cases. The training level of GC is higher than that of Basic, indicating that the adjacent word information extracted by the model is beneficial to model training. ATGC's training level is \revise{the highest}, indicating that the \revise{adversarial} perturbation is useful for model training.

	\begin{figure}
		\centering  
		\subfigure[The curves of training loss.]{
			\label{fig_train_loss}
			\includegraphics[width=0.8\textwidth]{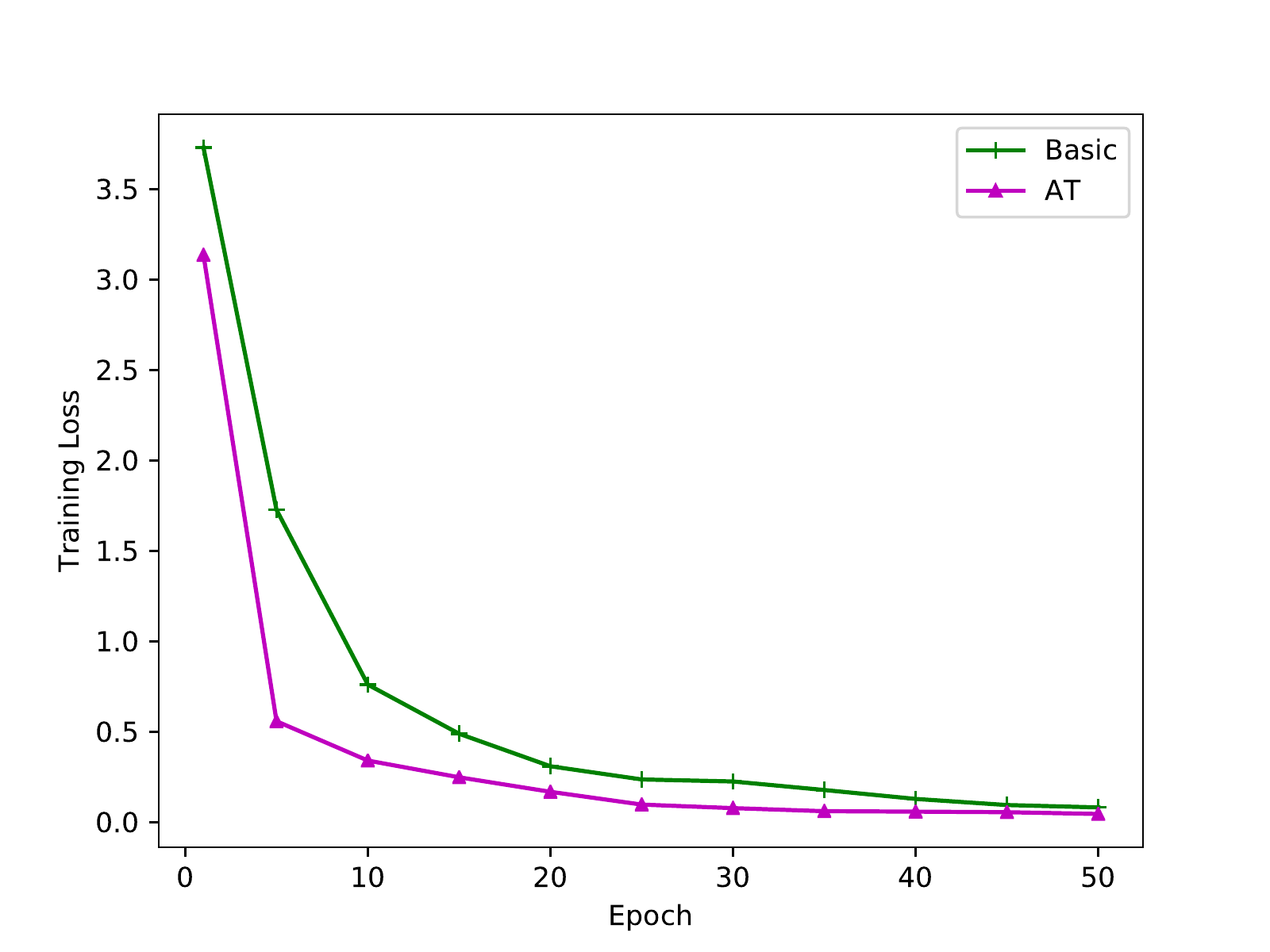}}
		\subfigure[The curves of Dev F1.]{
			\label{fig_dev_f1}
			\includegraphics[width=0.8\textwidth]{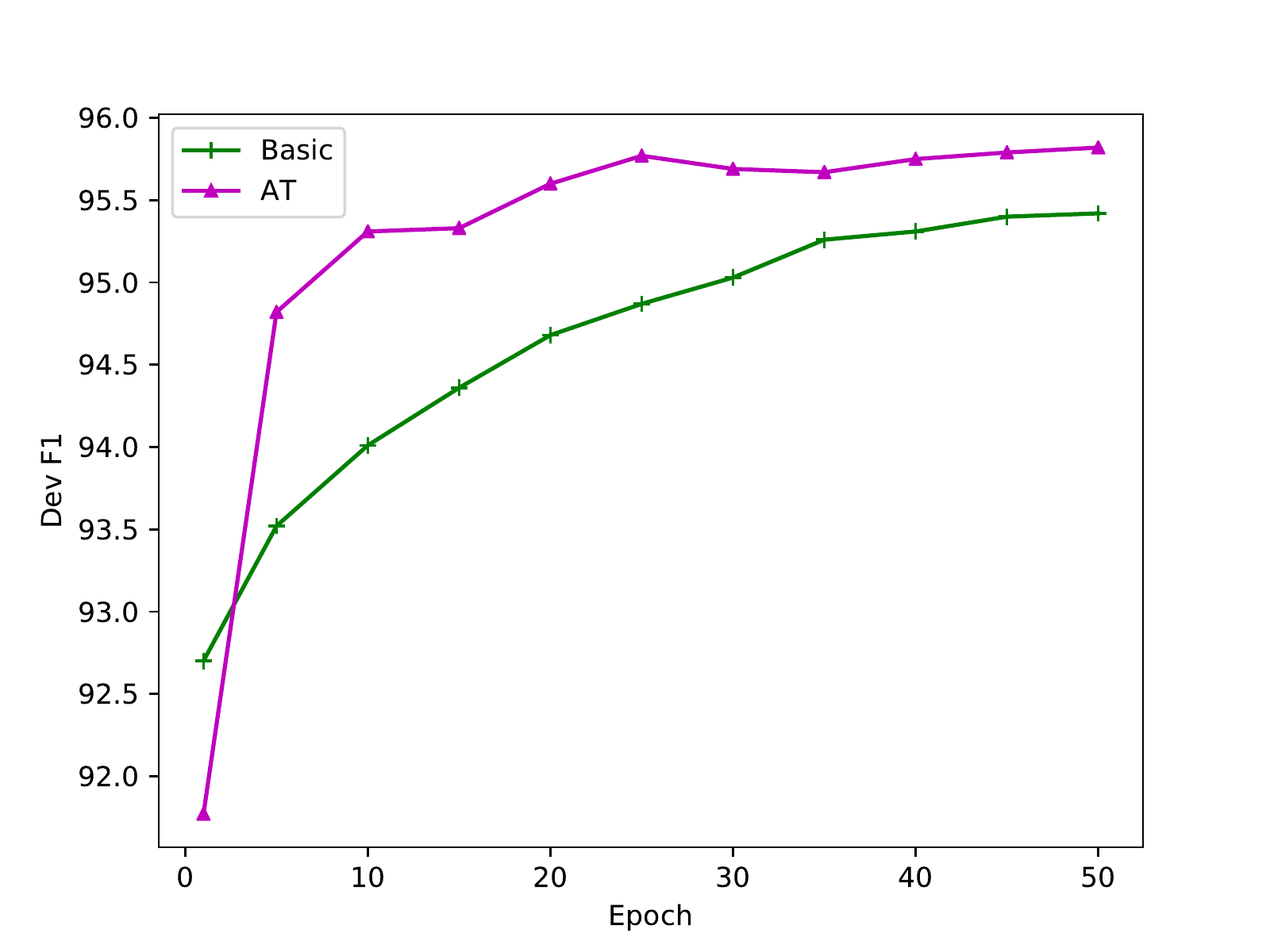}}
		\caption{The indicators of the training process with and without adversarial training (AT).}
		\label{fig_at}
	\end{figure}

	\subsection{Effect of Adversarial Training}
		
	Now we explore the \revise{effect} of adversarial training by presenting the indicators in the training process with and without it. \revise{Figure~\ref{fig_at} shows} the change of training loss and Dev F1 as the training epoch increases. \revise{We record the first 50 epochs to observe the situation during training.} The green curve represents the basic condition, and the carmine curve represents the AT condition.
	Figure~\ref{fig_train_loss} shows that the training loss of AT condition is lower, and the convergence speed is faster during training, \revise{especially in the first 30 epochs}. And the final training loss values of basic and AT condition \revise{are both close to 0.06} since they are both overfitting at that time. From Figure~\ref{fig_dev_f1}, it can be seen that the Dev F1 of AT condition increased faster, and its final value is higher. It indicates that adversarial training has an inhibitory effect on overfitting.
	
	\subsection{Case Study}
	
	\begin{table*} 
		\caption{Case study for the three datasets. Here ``\textit{LACK}'', ``\textit{WRONG}'', and ``\textit{CORRECT}'' indicate the meaning of absence, misclassification, and entirely correct respectively.}\label{tab_case_study}
		\centering
		\large
		\resizebox{\textwidth}{60mm}{
			\begin{tabular}{|c|p{4.2cm}|p{3.2cm}|p{3.2cm}|p{3.2cm}|p{3.2cm}|}
				\hline
				\multirow{2}*{\textbf{Dataset}}&\centering \multirow{2}*{\textbf{Sentence}} & \multicolumn{4}{c|}{\textbf{Named Entity}}\\
				\cline{3-6}
				
				&  &\centering  \textbf{GroundTruth}& \centering \textbf{Basic} &\centering \textbf{GC} &\centering \textbf{ATGC} \tabularnewline
				\hline
				\multirow{8}*{\textbf{CoNLL-03}} & Hosts UAE play Kuwait and South Korea take on Indonesia on Saturday in Group A matches. &LOC: UAE, Kuwait, South Korea, Indonesia& \textit{LACK} - LOC: Indonesia& \textit{CORRECT} &  \textit{CORRECT} \\
				\cline{2-6}
				&Top-seeded Eyles now meets titleholder Peter Nicol of Scotland who overcame Simon Parke of England. &PER: Eyles, Peter Nicol, Simon Parke; LOC: Scotland, England &\textit{LACK} - PER: Peter Nicol & \textit{WRONG} - LOC: Peter Nicol &\textit{CORRECT} \\
				\hline
				\multirow{6}*{\textbf{OntoNote 5.0}} &The same toy is sold for less than 40 US dollars at Wal-Mart. &MONEY: 40 US dollars; ORG: Wal-Mart & \textit{LACK} - MONEY: 40 US dollars& \textit{LACK} - MONEY: 40 US dollars&  \textit{CORRECT}\\
				\cline{2-6}
				&Last week's real Jackson story ran in the New York Daily News. &DATA: Last week; PERSON: Jackson; ORG: the New York Daily News & \textit{WRONG} - GPE: New York & \textit{CORRECT}& \textit{CORRECT}\\
				\hline
				\multirow{10}*{\textbf{WNUT-17}} &I will nominate Virgin Active at Moore Park / Zetland for you.  &corporation: Virgin Active; location: Moore Park, Zetland &\textit{CORRECT} &\textit{CORRECT} & \textit{CORRECT} \\
				\cline{2-6}
				&Why were Olive and Emma's powers changed in Miss Peregrint's Home for Peculiar Children? & person: Olive, Emma; creative-work: Miss Peregrint's Home, Peculiar Children;& \textit{WRONG} - person: Peregrint; \textit{LACK} - creative-work: Miss Peregrint's Home for Peculiar Children; &\textit{WRONG} - person: Peregrint; \textit{LACK} - creative-work: Miss Peregrint's Home for Peculiar Children; & \textit{LACK} - creative-work: Miss Peregrint's Home for Peculiar Children;\\
				\hline
			\end{tabular}
		}
		
	\end{table*}
	\begin{table} 
		\caption{\revise{An example of predicted probability distribution under different model conditions. The example is the first case of CoNLL-03 in Table~\ref{tab_case_study}, and these probability values are from the output of CRF Module. In this table, the darker background color of tokens means the higher probability of being predicted as LOC.}}\label{tab_GC_study}
		\centering
		\footnotesize
		\begin{tabular}{|c|c|c|}
			\hline
			\multicolumn{2}{|c|}{\textbf{Model}} & \textbf{Predicted probability of LOC}\\
			\hline
			\multirow{4}*{ASTRAL} & Basic&  \includegraphics[width=0.72\textwidth]{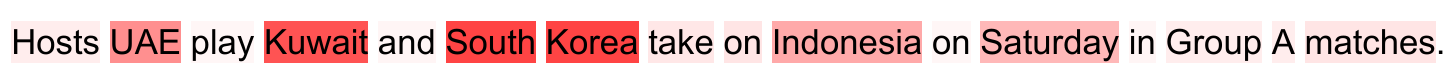}\\
			\cline{2-3}
			&GC &  \includegraphics[width=0.72\textwidth]{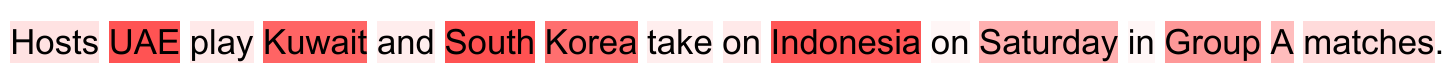}\\
			\cline{2-3}
			&ATGC &  \includegraphics[width=0.72\textwidth]{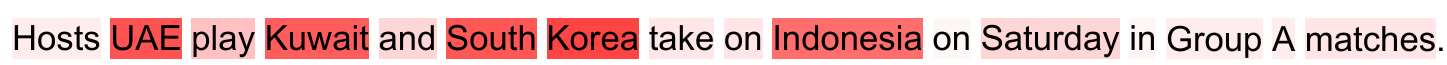}\\
			\hline
		\end{tabular}
	\end{table}

	We show several cases in Table~\ref{tab_case_study}. Two sentences from each dataset are selected to analyze the characteristics of the datasets and the changes in model results under different conditions. Here we choose the sentences with concentrated named entities to analyze the model performance of different conditions. The column of Ground Truth shows the standard answers. In the following three columns, Basic, GC, and ATGC, we list the differences between the corresponding model and ground truth. \revise{Here} ``LACK'', ``WRONG'', and ``CORRECT'' \revise{respectively} indicate the meaning of absence, misclassification, and entirely correct. We still use the given label form for each dataset, so that different datasets have different kinds of labels. For example, the geographically named entity labeled ``LOC'' in  CoNLL-03 is similar to ``location'' in WNUT, as well as ``GPE'' in OntoNote 5.0.
	
	Table~\ref{tab_case_study} indicates that the results of Basic, GC, and ATGC are getting better and better for these samples, which is consistent with the previous statistical results. From some examples, we notice that GC is benefiting from the adjacent words. In the first sentence of CoNLL-03, thanks to the help of ``on'', GC can solve the LACK of ``Indonesia''. In the second sentence of OntoNote 5.0, GC correctly identifies ``New York Daily News'' as an organization instead of recognizing ``New York'' itself as a location. \revise{We further analysis the first case in Table~\ref{tab_case_study} to show the actual impact of the model condition in terms of word choice. As shown in Table~\ref{tab_GC_study}, the darker the word's background in this table, the more likely the model recognizes it as LOC. Compared with Basic's result, GC's attention to `` Indonesia '' has increased significantly, but words such as `` Saturday '' and `` Group '' have also caused more interference at the same time. And the ATGC effectively suppresses this interference. In order to further explore the advantages of GC, we observe 50 cases per each dataset in which location entities are misclassified by GC though their adjacent words contain prepositions like ``on'', ``at''. We find that the percentages of these location entities which correctly identified under GC are 64\%, 56\%, and 68\% for CoNLL-03, OntoNote 5.0 and WNUT-17 respectively. This shows that GC can effectively reduce errors in these cases.} For ATGC, the named entities in the samples are almost extracted correctly. Benefiting from the adversarial training, ATGC can correctly recognize a rare name ``Peter Nicol'' as ``person'' instead of ``location''. Overall, the model has strong extraction capabilities for simple locations and organizational structures. However, specific words that require background knowledge, such as ``Miss Peregrint's Home for Peculiar Children'', are still hard to be extracted.

	\section{Conclusion and Future Work}
	In this paper, a NER system named ASTRAL is proposed, whose model structure and training process are augmented. We incorporated a Gated-CNN module with the network, helping the model to extract spatial information between adjacent words. In the training process, normalized adversarial training is introduced to enhance the model's robustness and generalization ability. We performed experiments on three benchmarks, and have shown that our system gets a significant improvement over previous work and achieves state-of-the-art performance.
	
	Our ASTRAL system has a notable performance on recognizing named entities from practical text, such as news, books, comments, etc. Thus this system could meet the requirement of users and advanced systems who need these named entities for further \revise{processing}. Compared to the recent research on the general language model such as ELMo~\cite{peters2018deep} and BERT~\cite{devlin2018bert}, our experiments show that stronger task-related modules could also have excellent effects. Meanwhile, the Gated-CNN and normalized adversarial training in this paper could be introduced into other neural language processing systems.
	
	In the future, we will mainly focus on the following two aspects. Firstly, the effect of different task-related modules combined with different language models is worth studying. Based on the characteristics of different language models, we will design matching task-related modules for each language model. Secondly, we will study the data enhancement methods, such as distant supervision, to solve the problem of insufficient training data. It is considered to be a direct means of solving the overfitting problem.
	
	\section*{References}
	
	\bibliography{mybibtex}
	
\end{document}